\documentclass[runningheads]{llncs}
\usepackage{graphicx}
\usepackage{subfigure}
\usepackage{comment}
\usepackage{amsmath,amssymb}
\usepackage{color}
\usepackage{bm}
\usepackage{multirow}
\usepackage{bbding}
\usepackage{epsfig}
\usepackage{eso-pic}

\begin{document}
\pagestyle{headings}
\mainmatter
\def\ECCVSubNumber{2822}

\title{Reinforced Axial Refinement Network for Monocular 3D Object Detection}
\titlerunning{Reinforced Axial Refinement Network for Monocular 3D Object Detection}
\author{Lijie Liu\inst{1} \and
Chufan Wu\inst{1}\and
Jiwen Lu\inst{1*}\and
Lingxi Xie\inst{2}\and
Jie Zhou\inst{1}\and
Qi Tian\inst{2}}
\authorrunning{L. Liu et al.}
\institute{Department of Automation, Tsinghua University, China\\
State Key Lab of Intelligent Technologies and Systems, China\\
Beijing National Research Center for Information Science and Technology, China \and
Huawei Inc.\\
\email{\{llj95luffy,chufanwu15,198808xc\}@gmail.com\\ \{lujiwen,jzhou\}@tsinghua.edu.cn, tian.qi1@huawei.com}}

\maketitle
\begin{abstract}
Monocular 3D object detection aims to extract the 3D position and properties of objects from a 2D input image. This is an ill-posed problem with a major difficulty lying in the information loss by depth-agnostic cameras. Conventional approaches sample 3D bounding boxes from the space and infer the relationship between the target object and each of them, however, the probability of effective samples is relatively small in the 3D space. To improve the efficiency of sampling, we propose to start with an initial prediction and refine it gradually towards the ground truth, with only one 3d parameter changed in each step. This requires designing a policy which gets a reward after several steps, and thus we adopt reinforcement learning to optimize it. The proposed framework, Reinforced Axial Refinement Network (RAR-Net), serves as a post-processing stage which can be freely integrated into existing monocular 3D detection methods, and improve the performance on the KITTI dataset with small extra computational costs.

\end{abstract}
\section{Introduction}

Over the past years, monocular 3D object detection has attracted increasing attentions in computer vision \cite{chabot2017deep,ku2019monocular,chang2019deep,roddick2018orthographic,simonelli2019disentangling}. For many practical applications such as autonomous driving \cite{bertozzi2000vision,geiger2012we,geiger2013vision,chen2015deepdriving,janai2017computer}, augmented reality \cite{alhaija2018augmented,rematas2018soccer} and robotic grasping \cite{saxena2008robotic,mahler2017dex,levine2018learning}, high-precision 3D perception of surrounding objects is an essential prerequisite. Compared to 2D object detection, monocular 3D object detection can provide more useful information including orientation, dimension, and 3D spatial location. However, due to the increase in dimensionality, the 3D Intersection-over-Union (3D-IoU) evaluation criterion is much more strict than 2D-IoU, making monocular 3D object detection a very difficult problem. In some challenging scenarios, state-of-the-art methods can only achieve a 3D average precision (3D AP) of around 10\% \cite{brazil2019m3d,ma2019accurate}.

\begin{figure}[t]
    \begin{center}
       \includegraphics[width=1\linewidth]{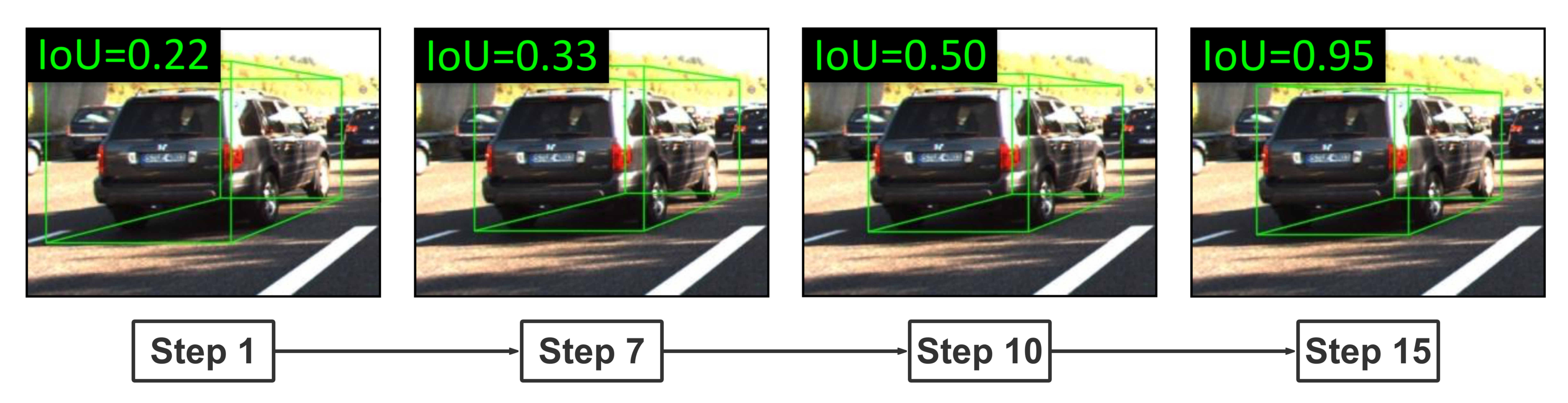}
    \end{center}
   \caption{Illustration of our idea that sequentially refines 3D detection using deep reinforcement learning. During the process, the 3D parameters are refined iteratively. In this example, we can see the trend that 3D-IoU gets improved as the 3D box gradually fits the object. Many intermediate steps are omitted here due to the limited space.}
\label{fig:idea}
\end{figure}

There have been a variety of efforts on detecting the objects in 3D space from a single image, and two popular trends are using geometry constraints \cite{mousavian20173d,kundu20183d,li2019gs3d} and depth estimation \cite{xu2018multi,qin2019monogrnet,manhardt2019roi,wang2019pseudo}. Due to the lack of real 3D cues, these methods often suffer from the problem of foreshortening (for distant objects, a tiny displacement on the image plane can lead to a large shift in the 3D space), and thus fail to achieve high 3D-IoU rates between detection results and ground-truth. To make up for the loss of 3D information, recently researchers propose to use a sampling-based method \cite{liu2019deep} to score the fitting degree between a sampled box and the object. However, in 3D space, the efficiency of sampling is very low and a randomly placed 3D box often has no overlap (3D-IoU is 0) to the target, which leads to inefficient learning. To this end, it is desirable to propose a method which can significantly increase the sampling efficiency.

In this paper, we ease this challenge by presenting a new framework called Reinforced Axial Refinement Network (RAR-Net), which, as illustrated in Fig.~\ref{fig:idea}, iteratively refines the detected 3D object to the most probable direction. In this way, the probability of effective sampling (finding a positive example with a non-zero 3D-IoU) increases with iteration. This is a Markov Decision Process (MDP), which involves optimizing a strategy that gets a reward after multiple steps. We train the model using a Reinforcement Learning (RL) algorithm.

RAR-Net takes the current status as input, and outputs one refining action at a time. In each step, to provide the current detection information as auxiliary cues, we project it to an image of the same spatial resolution as the input image (each face of the box is painted in a specific color), concatenate this additional image to the original input, and feed the 6-channel input to the RAR-Net. This implicit way of embedding the 2D image and 3D information into the same feature space brings consistent accuracy gain. Overall, RAR-Net is optimized smoothly during training, in particular, with the help of abundant training data that are easily generated by simply jittering the ground-truth 3D box. 

We conduct extensive experiments on the KITTI object orientation estimation benchmark, 3D object detection benchmark and bird's eye view benchmark. As a refinement step, RAR-Net works well upon four popular 3D detection baselines, improving the base detection accuracy by a large margin, while requiring relatively small extra computational costs. This implies its potential in real-world scenarios. In summary, our contributions are three-fold:
\begin{itemize}
\item To the best of our knowledge, this is the first work that applies deep RL to refine 3D parameters in an iterative manner.
\item We define the action space and state representation, and propose a data enhancement which embeds axial information and image contents.
\item RAR-Net is a plug-and-play refinement module. Experimental results on the KITTI dataset demonstrate its effectiveness and efficiency.
\end{itemize}

\section{Related Work}
\noindent
\textbf{Monocular 3D Object Detection.}
Monocular 3D object detection aims to generate 3D bounding-boxes for objects from single RGB images. It is more challenging than 2D object detection due to the increased dimension and the absence of depth information. Early studies use handcrafted approaches trying to design efficient features for certain domain scenarios \cite{payet2011contours,fidler20123d,pepik2015multi,chen2016monocular}. However, they suffer with the ability to generalize. Recently, researchers have developed deep learning based approaches aiming to solve this problem leveraging largely labeled data. One cut-in point is to use geometry constraints to make up for the lack of 3D information. Mousavian~\emph{et al.}~\cite{mousavian20173d} present MultiBin architecture for orientation regression and compute the 3D translation using tight constraint. Kundu~\emph{et al.}~\cite{kundu20183d} propose a differentiable Render-and-Compare loss to supervise 3D parameters learning. Li~\emph{et al.}~\cite{li2019gs3d} utilize surface features to explore the 3D structure information of the object. Apart from these pure geometry-based methods, there are some other methods which turn to the depth estimation to recover 3D information. One straightforward way is to first predict the depth map using the depth estimation module and then perform 3D detection using the estimated 3D depth~\cite{xu2018multi,manhardt2019roi,wang2019pseudo,ma2019accurate}. Another way is to infer instance depth instead of global depth map~\cite{qin2019monogrnet}, which does not require additional training data. Recently, Liu~\emph{et al.}~\cite{liu2019deep} propose to sample 3D bounding boxes from the space and introduce fitting degree to score the candidates. Brazil~\emph{et al.} \cite{brazil2019m3d} design a 3D region proposal network called M3D-RPN to generate 3D object proposals in the space. However, the performance of these methods is still limited because of the low efficiency of sampling in the 3D space. Our work jumps out of the limitation of trending object detection modules by iteratively refining the box to the ground-truth. It greatly solves the issue when network cannot directly regress to the goal detection and achieves better result.

\noindent
\textbf{Pose Refinement Methods.}
Our method belongs to the large category of coarse-to-fine learning \cite{cao2015look,yoo2015attentionnet,yu2018recurrent}, which refines visual recognition in an iterative manner. The approaches most relevant to ours are the iterative 3D object pose refinement approaches in \cite{manhardt2018deep,li2018deepim}. Manhardt~\emph{et al.}~\cite{manhardt2018deep} train a deep neural network to predict a translational and rotational update for 6D model tracking. DeepIM \cite{li2018deepim} aims to iteratively refine estimated 6D pose of objects given the initial pose estimation. They also see the limitation of direct regression of images. However, these methods require the CAD model of the objects for fine correction and cannot be used in autonomous driving directly. In our case, we do not require complex CAD models and optimize the whole pose refinement process using deep RL.

\noindent
\textbf{Deep RL.}
RL aims at maximizing a reward signal instead of trying to generate a representational hidden state like traditional supervised learning problem \cite{littman2015reinforcement,mnih2015human,sutton2018reinforcement}.
Deep RL is the method of incorporating RL with deep learning. Due to the distinguished feature of delayed reward and the massive power of deep learning, deep RL has been widely used on decision making in goal-oriented problems like object detection \cite{caicedo2015active,mathe2016reinforcement}, deformable face tracking \cite{guo2018dual}, interaction mining \cite{duan2018graphbit}, object tracking \cite{yun2017action,ren2018deep} and video face recognition \cite{rao2017attention}. However, to our best knowledge, little work has been made in RL for pose refinement, especially in monocular 3D object detection. Our approach sees 3D parameter refinement problem as a multi-step decision-making problem by updating the 3D box using action from each step, which takes advantage of trial-and-error search in RL to achieve better result.

\section{Approach}

The monocular 3D object detection task requires solving a 9-Degree-of-Freedom (9-DoF) problem including dimension, orientation and location using a single RGB image as input. In this paper, we focus on improving the detection accuracy in the context of autonomous driving, where the object can only rotate around the Y axis, so the orientation has only 1-DoF. Although many excellent methods have been proposed so far, the monocular 3D object detection accuracy is still below satisfactory. So, we formulate the refinement problem as follows: given an initial estimation $(\hat{x}, \hat{y}, \hat{z}, \hat{h}, \hat{w}, \hat{l},\hat{\theta})$, the refinement model predicts a set of displacement values $(\Delta x, \Delta y, \Delta z, \Delta h, \Delta w, \Delta l, \Delta \theta)$. Then, a new estimation is computed as $(\hat{x}+\Delta x, \hat{y}+\Delta y, \hat{z}+\Delta z, \hat{h}+\Delta h, \hat{w}+\Delta w, \hat{l}+\Delta l, \hat{\theta}+\Delta \theta)$ and fed into the refinement model again. After several iterations, the refinement model can generate more and more accurate estimates.

\subsection{Baseline and the Curse of Sampling in 3D Space}
Monocular 3D object detection is an ill-posed problem, \textit{i.e.}, to recover 3D perception from 2D data. Although some powerful models have been proposed for 3D understanding \cite{mousavian20173d,qin2019monogrnet,brazil2019m3d}, it is still difficult to build relationship between the depth-agnostic input image and the desired 3D location. To alleviate the information gap, researchers came up with an alternative idea that samples a number of 3D boxes from the space and asks the model to judge the IoU between the target object and each sampled box \cite{liu2019deep}. Such models, sometimes referred to as a fitting network, produced significant improvement under sufficient training data and the help of extra (\textit{e.g.}, geometric) constraints.

However, we point out that the above sampling-based approaches suffer a difficulty in finding `effective samples' (those having non-zero overlap with the target) especially in the testing stage. This is mainly caused by the increased dimensionality: the probability that a randomly placed 3D box has overlap to a pre-defined object is much lower than that in the 2D scenario. For example, if we use a Gaussian distribution with a deviation of 1 meter, there is only a chance of 0.12 to place an effective sample on a car that is 5 meters away from the initial detection result. This situation even deteriorates with the distance becomes larger. That being said, unless the initial detection is sufficiently accurate, the sampling efficiency can be very low.

\subsection{Towards Higher Sampling Efficiency}

To improve the sampling efficiency, a straightforward idea is to go towards a roughly correct direction and then perform sampling at a better place. For the same example of the car that is 5 meters behind the detection result, if we move the current detection result towards the back direction for 2 meters, the possibility of sampling a non-zero IoU box will increase to 0.63. Furthermore, with multi-step refinement, the 3D box can even converge to the ground-truth and sampling becomes unnecessary. 

There are many moving options to choose, and we find that moving in only one direction at a time is the most efficient, because the training data collected in this way is the most concentrated (the output targets will not be scattered throughout the three-dimensional space). Most existing refinement models choose to optimize their objective function using one-step optimization, which learns to move from the initial estimate to the ground-truth directly. However, one-step optimization can barely achieve the global optimum, especially when there is more than one variable to be refined, because different variables can have an effect on each other. For example, refining the orientation first can help the model make better use of appearance information to refine to a more precise location. Two-stage cascaded refinement algorithm is another design choice, but it may bring considerable difficulties in algorithm design, especially in the way of defining different stages. Also, it is a challenging topic to prepare data for each stage, e.g. how to guarantee the training input fed into the second stage match the case in testing scenario.

Motivated by this concern, we choose to optimize the learning objective for the entire MDP instead of one step using RL-based framework which can support an arbitrary number of stages and the training procedure is elegant (few heuristic rules are required). Our approach starts from an initial estimate $(\hat{x}, \hat{y}, \hat{z}, \hat{h}, \hat{w}, \hat{l},\hat{\theta})$, and outputs a refining operation at a time. The 3D-IoU of the predicted object is therefore improved as the refinement of the 3D parameters.

\begin{figure}[t]
    \begin{center}
       \includegraphics[width=1\linewidth]{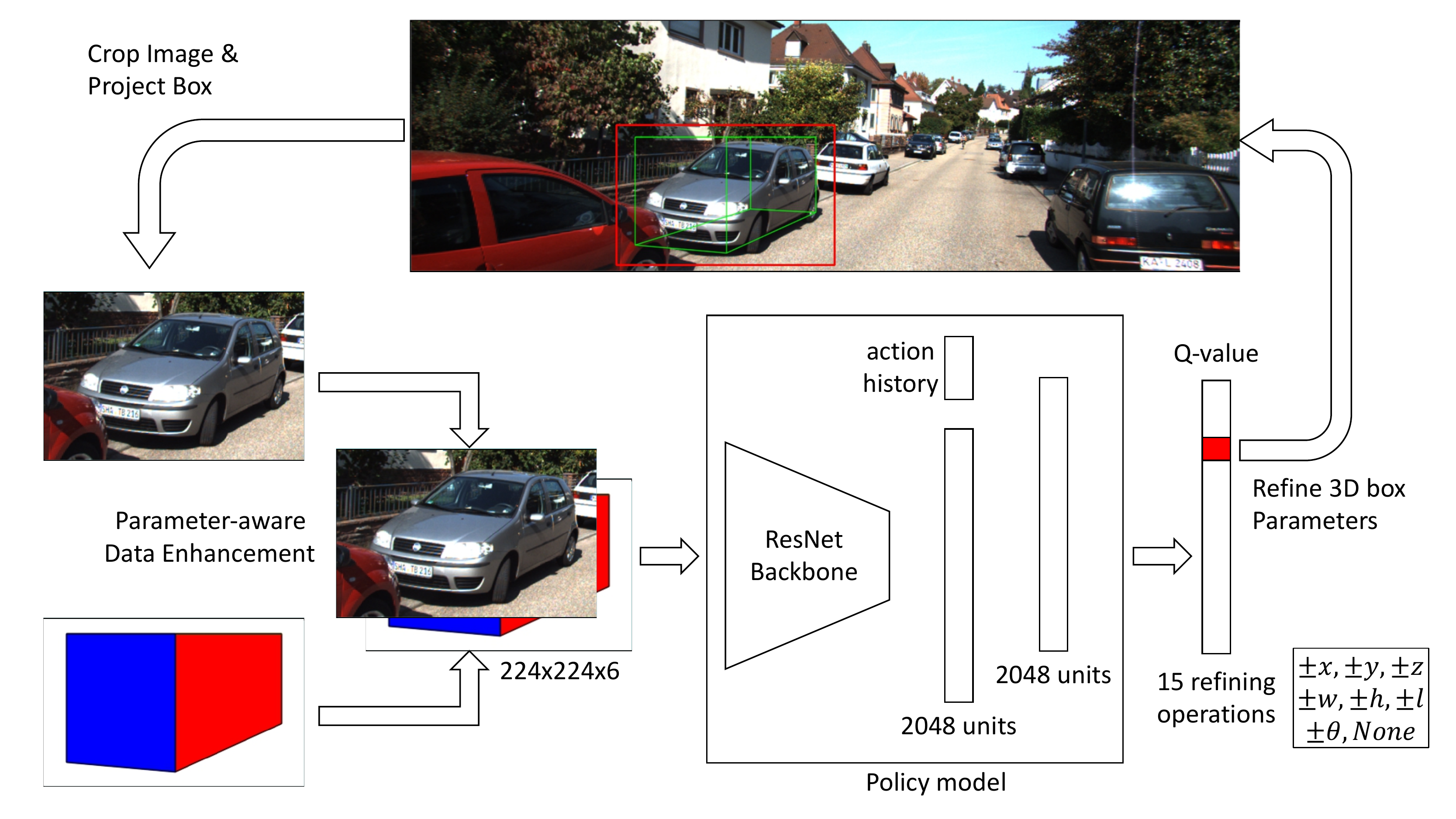}
    \end{center}
   \caption{The proposed framework for monocular 3D object detection. It is an iterative algorithm optimized by RL. In each iteration, an input image is enhanced by a parameter-aware mask and fed into a deep network, which produces a Q-value for each action as output and the 3D box is refined according to $\epsilon$-greedy policy.}
\label{fig:framework}
\end{figure}

Fig.~\ref{fig:framework} shows our overall pipeline, Reinforced Axial Refinement Network (RAR-Net), where we first enhance the input information using a parameter-aware module and then use a ResNet-101 \cite{he2016deep} backbone to output the action value (Q-value). Similar to \cite{caicedo2015active}, we also use the history vector to encode 10 past actions in order to stabilize search trajectories that might get stuck in repetitive cycles. We formulate the process of refining the 3D box from initial coarse estimate to the destination as an MDP and introduce an RL method for optimization. The goal is to predict a tight bounding-box with a high 3D-IoU.

\subsection{Refining 3D Detection with Reinforcement Learning}

In the RL setting, the optimal policy of selecting actions should maximize the sum of expected rewards $\mathbf{R}$ on a given initial estimated state $\mathbf{S}_i$. Since we do not have a priori knowledge about the optimal path to refine the initial predicted 3D bounding-box to the destination, we address the learning problem through standard DQN \cite{mnih2015human}. This approach learns an approximate action value function $Q(\mathbf{S}_i, \mathbf{A}_i)$ for each action $\mathbf{A}_i$, and selects the action with the maximum value as the next action to be done at each iteration. In order to prevent falling into local optimum, we use $\epsilon$-greedy policy, where there is certain possibility to choose random actions. The learning process iteratively updates the action-selection policy by minimizing the following loss function:
\begin{equation}
\mathcal{L}(\theta) =  [ \mathbf{R}_i + \gamma \max_{\mathbf{A}_{i+1}}Q(\mathbf{S}_{i+1}, \mathbf{A}_{i+1};\theta^{-1}) -Q(\mathbf{S}_i, \mathbf{A}_i;\theta)  ]^2,
\end{equation}
where $\gamma$ is the discount factor, $\theta$ are the parameters of the Q-network, and $\theta^{-1}$ are the parameters of the target-Q-network, whose weights are kept frozen most of the time, but are updated with the Q-network's weights every few hundred iterations. We use $[\mathbf{R}_i + \gamma \max_{\mathbf{A}_{i+1}}Q(\mathbf{S}_{i+1}, \mathbf{A}_{i+1};\theta^{-1})]$ to approximate the optimal target value, because the optimal action-value function obeys the Bellman equation:
\begin{equation}
Q^{\star}(\mathbf{S}_i, \mathbf{A}_i) =  \mathbb{E}_{\mathbf{S}_{i+1}}[ \mathbf{R}_i + \gamma \max_{\mathbf{A}_{i+1}}Q^{\star}(\mathbf{S}_{i+1}, \mathbf{A}_{i+1})|\mathbf{S}_i, \mathbf{A}_i].
\end{equation}

Under our refinement problem setting, for the output Q-value, we use a 15-dimensional vector to represent 15 different refining operations and actions are chosen based on $\epsilon$-greedy policy. Considering that continuous action space is too large and difficult to learn, we set the refinement value to be discrete during each iteration. In practice, we define the refinement value as a fixed ratio of the corresponding dimension of the object. We present the detailed settings of the definition of state, action, state transition and reward of our refinement framework for monocular 3D object detection as follows:

\noindent
\textbf{State:} In this work, we define the state to include both the observation image patch and the projected 3D cuboid. Given an initial estimate of the object $\mathbf{X} = (\hat{x}, \hat{y}, \hat{z}, \hat{h}, \hat{w}, \hat{l},\hat{\theta})$, which is often the detection results of other monocular 3D object detection methods, we use a standard camera projection  to obtain the top left point and bottom right point of the crop image patch:
\begin{equation}
(u_{\min}, v_{\min}, u_{\max}, v_{\max}) = \mu(\mathbf{X}, \mathbf{K}),
\end{equation}
where $\mathbf{K} \in \mathbb{R}^{3\times4}$ is the camera intrinsic matrix and the function $\mu$ is the projection operation.
To include more context information, we enlarge the patch regions by a factor of 1.2 in height and width. For the projected 3D cuboid, we crop in the same position as the image patch and use white color as the background. Therefore, our state is a 6-channal image patch:
\begin{equation}
\mathbf{S} = [\phi(u_{\min}, v_{\min}, u_{\max}, v_{\max}, \mathbf{I}); \mathbf{P}(\mathbf{X}, \mathbf{K})],
\end{equation}
where $\mathbf{I}$ is the original image, $\mathbf{P}(\mathbf{X}, \mathbf{K})$ is the projected 3D cuboid and $\phi\!\left(\cdot\right)$ is the crop operation. Finally, $\mathbf{S}$ is resized to fit the input size of RAR-Net.
 
\begin{figure}[t]
\centering
\subfigure[World Coordinate System]
{\includegraphics[width=0.49\textwidth]{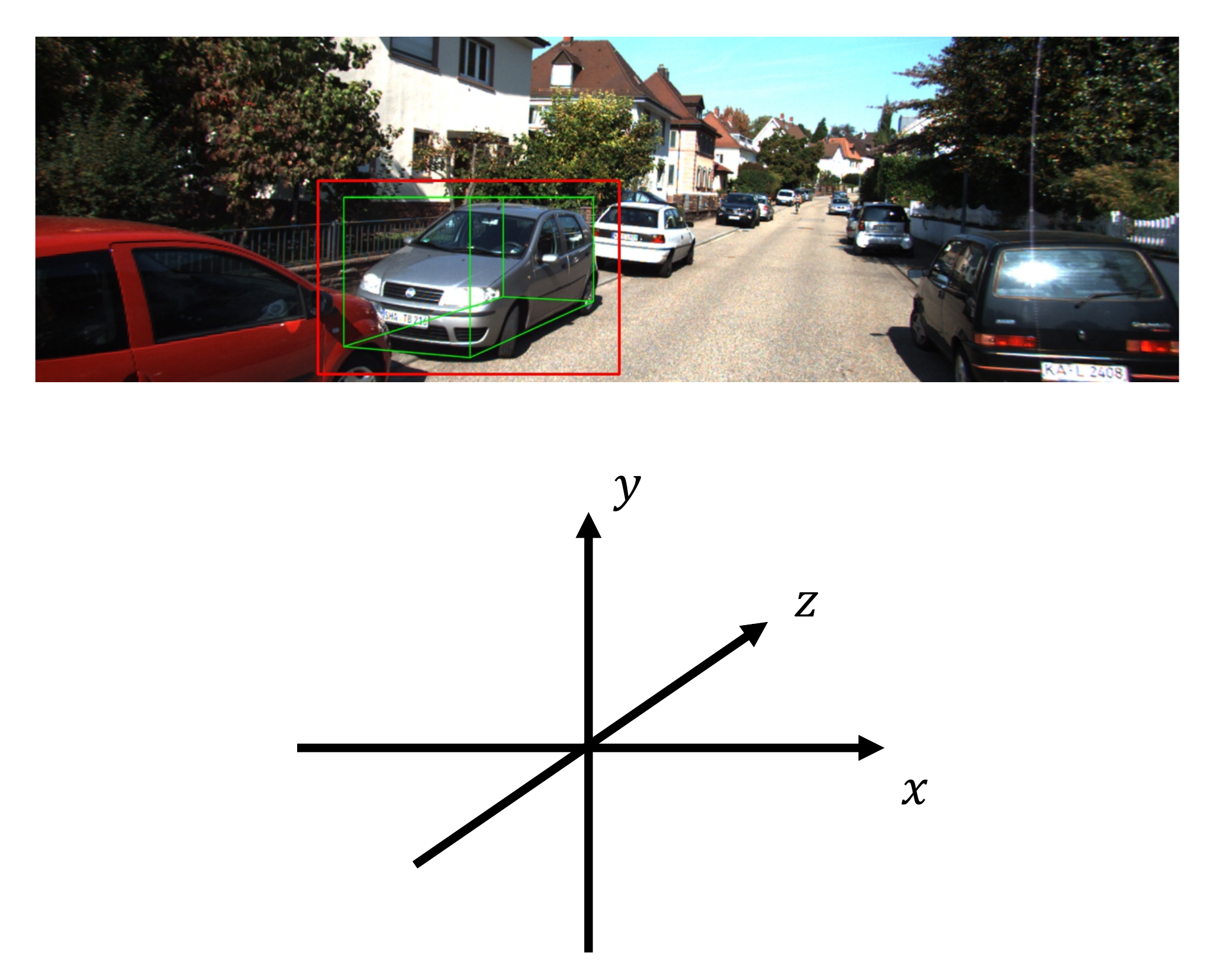}}
\subfigure[Axial Coordinate System]
{\includegraphics[width=0.49\textwidth]{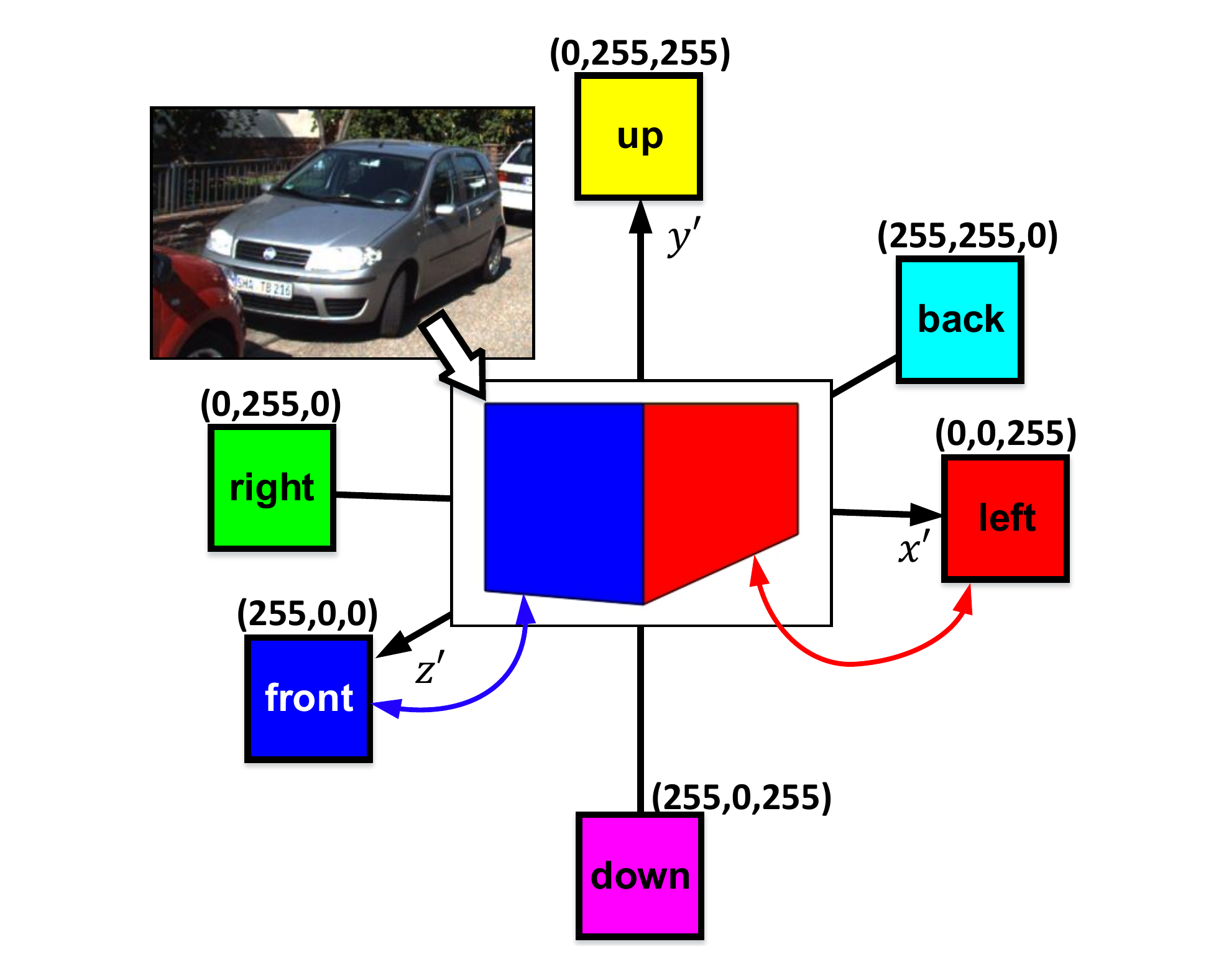}}
\caption{(a) shows the world coordinate system, which is related to the camera pose and shared by all the objects. (b) shows the axial coordinate system for one sample object. We also illustrate how to generate the parameter-aware mask from a 3D object (best viewed in color). Each color indicates one fixed face. Only two faces are visible in this real example.}
\label{fig:sample}
\end{figure}

\noindent
\textbf{Action:} Our action set $\mathcal{A}$ consists of 15 refining operations, including a none operation indicating no refinement. These operations are related to the 3D parameters of the detections. For instance, the action $+\Delta x$ will lead to a displacement along the width axis of the object with the value $(\Delta x' = \delta \times \hat{w})$, where $\delta$ is a fixed ratio. It is worth mentioning that there are two choices for the definition of our shifting actions, one is defined in the world coordinate system and the other is defined in the axial coordinate system of the object as shown in Fig.~\ref{fig:sample}. If we need to move the object to the left in the world coordinate system, for the former definition, we have to predict the same moving action for cars with different orientations (appearances). But, if we use the latter definition, the shifting operation will be related to the orientation of the cars, thus turning a many-to-one mapping to one-to-one mapping and easing the training process. 

\noindent
\textbf{State Transition:} Our state transition function ${T}$ refines the predicted box of the objects from $\mathbf{X}_i= (\hat{x}, \hat{y}, \hat{z}, \hat{h}, \hat{w}, \hat{l},\hat{\theta})$ to $\mathbf{X}_{i+1}= (\hat{x}+\Delta x, \hat{y}+\Delta y, \hat{z}+\Delta z, \hat{h}+\Delta h, \hat{w}+\Delta w, \hat{l}+\Delta l, \hat{\theta}+\Delta \theta)$. However, the moving direction is defined along the coordinate axes of the object, while the $(\hat{x}, \hat{y}, \hat{z})$ is defined in the world coordinate system, so we need to transform the displacement value across two different coordinate systems. Denote the output displacement of RAR-Net as $(\Delta x', \Delta y', \Delta z')$, which is defined in the axial coordinate system, we have:
\begin{equation}
\left\{
\begin{aligned}
    	\Delta x =& \Delta z' \times \cos\hat{\theta} +  \Delta x' \times \sin\hat{\theta}\\
	\Delta y =& \Delta y'   \\
	\Delta z =& \Delta z' \times (-\sin\hat{\theta}) +  \Delta x' \times \cos\hat{\theta}
\end{aligned}
\right.
\end{equation}
Therefore, we can translate state $\mathbf{S}_i$ to state $\mathbf{S}_{i+1}$ according to the output displacement value of RAR-Net.

\noindent
\textbf{Reward:} The reward function ${R}$ reflects the detection accuracy improvement from state $\mathbf{S}_i$ to $\mathbf{S}_{i+1}$. Considering that increasing the 3D-IoU will have positive reward and decreasing the 3D-IoU will have negative reward, we define the reward function as:
\begin{equation}
\mathbf{R}_i=
\left\{
\begin{aligned}
    	&+1,&\mathrm{if}\;\Delta \mathrm{IoU}_\mathrm{3D} > 0\\
	&-1,&\mathrm{if}\;\Delta \mathrm{IoU}_\mathrm{3D} < 0\\
	&\mathrm{sgn}[(\mathbf{X}_{i+1}-\mathbf{X}_i)({\mathbf{X}}^{\star}-\mathbf{X}_i)],&\mathrm{if}\;\Delta \mathrm{IoU}_\mathrm{3D} = 0\\
\end{aligned}
\right.
\end{equation}
where ${\mathbf{X}}^{\star}$ is the ground-truth 3D parameters and $\Delta \mathrm{IoU}_\mathrm{3D}$ is the changes of 3D-IoU. When there is no overlap between the estimated and ground-truth boxes, we use the changes in 3D parameters as the reward signal. In addition, when we arrive at a none action or the end of the sequence, we set the reward to $+3$ for a successful refinement (IoU $\geq$ 0.7), and $-3$ otherwise.

\subsection{Parameter-aware Data Enhancement}

In our iteration-based framework, two input sources are necessary, namely, an image patch which lies in the 2D image space (high-level image features), and the current detection result which lies in the 3D physical space (low-level geometry features). Provided that the desired output is strongly related to both information, it remains unclear how to combine both cues, in particular they come from two domains which are quite different from each other. 
Based on the above motivation, we propose to attach the refined result of last iteration into the input of current iteration. There are many options to achieve this goal, and the naive case is to concatenate the 3D parameters and the image feature in a late-fusion manner, but this practice can barely provide enough appearance cues. Another way is to project the 3D bounding-box on the input image patch \cite{liu2019deep} or render the 3D object when 3D CAD models are available \cite{kundu20183d}, but these methods may damage the original information since the projection result will obscure the original image. 

To avoid loss of information while providing sufficient appearance cues, we propose to project the 3D bounding box on the 2D image plane, and draw different colors on each face of the projected cuboid. This idea is similar to~\cite{ren2018deep}. In order to prevent loss of depth information during the projection operation, we embed the instance depth into the intensity of the color as $c'$, where $c'=c \times \frac{128}{255}$ if $z > 50$, and $c'=c \times (1-\frac{z}{100})$ if $z \le 50$, $c$ is the base RGB value shown in Fig.~\ref{fig:sample}, and $z$ is the instance depth of the object. Thus, different appearance will represent different 3D parameters of the object. For example, we paint blue for the front face, so the blue cue can guide the model to learn the refining policy along the forward-backward axis. A sample projection is shown in Fig.~\ref{fig:sample}. We concatenate the painted cuboid and the original image patch to construct a 6-channel input patch as the final input of our RAR-Net.

For the painting process, we use the OpenCV function {\tt{fillConvexPoly}} to color each face of the projected cuboid. We also apply black to the edges of the projected cuboid to strengthen the boundary. Since some faces are invisible from the front view, we have to determine the visibility of each face. Denote the center of $i$-th face as $\mathbf{C}_i$, and the center of the 3D bounding box as $\mathbf{C}$, the visibility of $i$-th face, $V_i$, is determined by whether $(\bm{0}- \bm{C})(\bm{C}_i-\bm{C})$ is greater than 0.

\subsection{Implementation Details}
\textbf{Training:} We used the ResNet-101 as backbone, and changed the input size into $224\times 224 \times 6$, and the output size into $15$. We trained the model from scratch. In order to speed up the RL process, we first performed supervised pre-training using one-step optimization where the model learns to perform the operation with the largest amount of correction. To create the training set, we added a jitter of Gaussian distribution to the 3D bounding boxes and each object leads to 300 training samples, whose projection is checked to be inside the image space. During the pre-training process, the model was trained with SGD optimizer using a start learning rate of $10^{-2}$ with a batch size of 64. The model was trained for 15 epochs and the learning rate was decayed by 10 every 5 epochs. During RL, The model was trained with Adam optimizer using a start learning rate of $10^{-4}$ with a batch size of 64 for 40000 iterations. We used memory replay \cite{schaul2015prioritized} with buffer size of $10^4$. The target Q-Network is updated for every 1000 iterations. The $\epsilon$ for greedy policy is set to 0.5 and will decay exponentially towards 0.05. The discount factor $\gamma$ is set to 0.9.
\noindent
\textbf{Testing:} we set the total refinement steps to 20, and during each step, we chose the action based on  $\epsilon$-greedy policy, which is  to take actions either randomly or with the highest action-value. For each action, the refining stride was set to $0.05\times$ corresponding dimensions. The $\epsilon$ for greedy policy is set to 0.05.

\section{Experiments}
\subsection{Dataset and Evaluation}
We evaluate our method on the real-world KITTI dataset \cite{geiger2012we}, including the object orientation estimation benchmark, the 3D object detection benchmark, and the bird's eye view benchmark. There are 7481 training images and 7518 testing images in the dataset, and in each image, the object is annotated with 2D location, dimension, 3D location, and orientation. However, only the labels in the KITTI training set are released, so we mainly conduct controlled experiments in the training set. Results are evaluated based on three levels of difficulty, namely, Easy, Moderate, and Hard, which are defined according to the minimum bounding-box height, occlusion, and truncation grade. There are two commonly used train/val experimental settings: Chen~\emph{et al.}~\cite{chen20153d,chen2016monocular} (val 1) and Xiang~\emph{et al.}~\cite{xiang2015data,xiang2017subcategory} (val 2). Both splits guarantee that images from the training set and validation set are sampled from different videos. 

We evaluate 3D object detection results using the official evaluation metrics from KITTI. 3D box evaluation is conducted on both two validation splits (different models are trained with the corresponding training sets). We focus our experiments on the car category as KITTI provides enough car instances for our method. Following the KITTI setting, we perform evaluation on the three difficulty regimes individually. In our evaluation, the 3D-IoU threshold is set to be 0.5 and 0.7 for better comparison. We compute the Average Orientation Similarity (AOS) for the object orientation estimation benchmark, the Average Precision (AP) for the bird's eye view boxes (which are obtained by projecting the 3D boxes to the ground plane), and the 3D Average Precision (3D AP) metric for evaluating the full 3D bounding-boxes.

\subsection{Comparison to the State-of-the-Arts}

To demonstrate that our proposed refinement method's effectiveness, we use the 3D detection results from different state-of-the-art 3D object detectors including Deep3DBox \cite{mousavian20173d}, MonoGRNet \cite{qin2019monogrnet}, GS3D \cite{li2019gs3d} and M3D-RPN \cite{brazil2019m3d} as the initial coarse estimates. These detection results are provided by the authors, except that we reproduce M3D-RPN by ourselves.

\begin{table}[t]
\caption{Comparisons of the Average Orientation Similarity (AOS, \%) to baseline methods on the KITTI orientation estimation benchmark. (In each group, we also show the 2D Average Precision (2D AP) of 2D detection results, which is the upper bound of AOS).}
\begin{center}
\scalebox{0.82}
{
\begin{tabular}{|c|c|c|c|c|c|c|}
\hline
\small \multirow{2}*{Method} &  \multicolumn{2}{c|}{Easy} & \multicolumn{2}{c|}{Moderate} & \multicolumn{2}{c|}{Hard}\\
\cline{2-7}
~ &  val 1 & val 2 & val 1 & val 2  &val 1 & val 2   \\
\hline\hline
Deep3DBox \cite{mousavian20173d} &- &98.59 (98.84)& - & \textbf{96.69} (97.20) & -  & 80.50 (81.16)\\
+RAR-Net &- & \textbf{98.61}  (98.84)&- &96.68 (97.20) & -& \textbf{80.51}  (81.16) \\
\hline
MonoGRNet \cite{qin2019monogrnet} &87.83 (88.17) &-& 77.80 (78.24)& -& 67.49 (68.02) & -\\
+RAR-Net& \textbf{87.86} (88.17) & - & 77.80  (78.24)& -&\textbf{67.51} (68.02)&-   \\
\hline
 GS3D \cite{li2019gs3d} &81.08 (82.02)&81.02 (81.66)& 73.01 (74.47) & 70.76 (71.68) & 64.65 (66.21)  & 61.77 (62.80)\\
+RAR-Net&\textbf{81.32} (82.02)& \textbf{81.21} (81.66)&\textbf{73.64} (74.47)&\textbf{70.92} (71.68)& \textbf{64.89} (66.21)&\textbf{61.88} (62.80)\\
\hline
M3D-RPN \cite{brazil2019m3d} &90.71 (91.49) &- &82.50 (84.09)& - & 66.44 (67.94) & -\\
+RAR-Net&\textbf{91.01} (91.49)& -&\textbf{82.92} (84.09)&- & \textbf{66.74} (67.94)&- \\
\hline
\end{tabular}
}
\end{center}
\label{tab:orientation}
\end{table}

\begin{table}[t]
\begin{center}
\centering
\caption{Comparisons of 3D localization accuracy (AP, \%) to state-of-the-arts methods on the KITTI bird's eye view benchmark.}
\scalebox{0.85}{
\begin{tabular}{|c|c|c|c|c|c|c|c|c|c|c|c|c|}
\hline
\multirow{3}*{Method} &  \multicolumn{6}{c|}{IoU = 0.5}&  \multicolumn{6}{c|}{IoU = 0.7}\\
\cline{2-13}
~ & \multicolumn{2}{c|}{ Easy} & \multicolumn{2}{c|}{Moderate} & \multicolumn{2}{c|}{Hard}  &\multicolumn{2}{c|}{ Easy} & \multicolumn{2}{c|}{Moderate} & \multicolumn{2}{c|}{Hard}  \\
\cline{2-13}
~ &  val 1 & val 2&  val 1 & val 2&  val 1 & val 2&  val 1 & val 2&  val 1 & val 2&  val 1 & val 2 \\
\hline\hline
Deep3DBox \cite{mousavian20173d} & - &30.02 & - &23.77 & -  & 18.83 &- & 9.99 & - &7.71 &- &5.30\\
+RAR-Net & - &\textbf{33.12} & - &\textbf{24.42} & -  & \textbf{19.11} &- &\textbf{14.38}& - &\textbf{10.28} &- &\textbf{8.29}\\
\hline
MonoGRNet \cite{qin2019monogrnet} &53.91 & - &39.45 & - &32.84  & - &24.84 & - &19.27 & - &16.20 & -\\
+RAR-Net&\textbf{54.01} & - &\textbf{41.29} & - &\textbf{32.89}  & - &\textbf{26.34} & - &\textbf{23.15} & - &\textbf{19.12} & -\\
\hline
 GS3D \cite{li2019gs3d} &38.24 &46.50 & 32.01 &39.15 & 28.71  & 33.46 &14.34 & 20.00 & 12.52 &16.44 & 11.36 &13.40\\
+RAR-Net &\textbf{38.31} &\textbf{48.90} & \textbf{34.01} &\textbf{39.91} & \textbf{29.70 } &\textbf{ 35.16} &\textbf{18.47} & \textbf{24.29} & \textbf{16.21} &\textbf{19.23}& \textbf{14.10} &\textbf{15.92}\\
\hline
M3D-RPN \cite{brazil2019m3d} &56.92&- &43.03 &-& 35.86  & - &27.56&- & 21.66 &-& 18.01  & -\\
+RAR-Net&\textbf{57.12}&- &\textbf{44.41} &-& \textbf{37.12}  & - &\textbf{29.16}&- & \textbf{22.14} &-& \textbf{18.78}  & -\\
\hline
\end{tabular}
}
\label{tab:locview}
\end{center}
\end{table}

We first compare AOS with these baseline methods, and the results are shown in Table \ref{tab:orientation}. The 2D Average Precision (2D AP) is the upper bound of AOS by definition, and we can see that our refinement method can improve the baseline even if the performance is already very close to the upper bound. Then we compare 2D AP in bird's view of our method with these published methods. As can be seen in Table \ref{tab:locview}, our method improve the existing monocular 3D object detection methods for a large margin. For example, the AP of Deep3DBox in the setting of $\mathrm{IoU}=0.7$ gains a $4\%$ improvement. We also notice that for different baselines, our improvements differ -- for the lower baseline, the improvements are larger because they have more less perfect detection results. Similarly, we report a performance boost on 3D AP as shown in Table \ref{tab:location}. In addition, our method works better in the hard scenario that requires $\mathrm{IoU}=0.7$.

\begin{table}[t]
\centering
\caption{Comparisons of 3D detection accuracy (AP, \%) with state-of-the-arts on the KITTI 3D object detection benchmark.}
\begin{center}
\scalebox{0.85}{
\begin{tabular}{|c|c|c|c|c|c|c|c|c|c|c|c|c|}
\hline
\multirow{3}*{Method} &  \multicolumn{6}{c|}{IoU = 0.5}&  \multicolumn{6}{c|}{IoU = 0.7}\\
\cline{2-13}
~ & \multicolumn{2}{c|}{ Easy} & \multicolumn{2}{c|}{Moderate} & \multicolumn{2}{c|}{Hard}  &\multicolumn{2}{c|}{ Easy} & \multicolumn{2}{c|}{Moderate} & \multicolumn{2}{c|}{Hard}  \\
\cline{2-13}
~ &  val 1 & val 2&  val 1 & val 2&  val 1 & val 2&  val 1 & val 2&  val 1 & val 2&  val 1 & val 2 \\
\hline\hline
Deep3DBox \cite{mousavian20173d} & - &27.04 & - &20.55 & -  & 15.88 &- & 5.85 & - &4.10 &- &3.84\\
+RAR-Net & - &\textbf{28.92} & - &\textbf{22.13} & -  &\textbf{ 16.12} &- & \textbf{14.25} & - &\textbf{9.90} &- &\textbf{6.14}\\
\hline
MonoGRNet \cite{qin2019monogrnet} &50.27 & - &36.67 & - &30.53  & - &13.84 & - &10.11 & - &7.59 & -\\
+RAR-Net&\textbf{54.17 }& - &\textbf{39.71} & - &\textbf{31.82}  & - &\textbf{18.25} & - &\textbf{14.40} & - &\textbf{11.98} & -\\
\hline
GS3D \cite{li2019gs3d} &30.60 &42.15&26.40 &31.98 & 22.89  & 30.91 &11.63 & 13.46 & 10.51 &10.97 & 10.51 &10.38\\
+RAR-Net &\textbf{33.12} &\textbf{42.29}&\textbf{28.11 }&\textbf{32.18 }& \textbf{24.12  }&\textbf{ 31.85} &\textbf{17.82 }& \textbf{19.10 }& \textbf{14.71 }&\textbf{15.72} & \textbf{14.81 }&\textbf{13.85}\\
\hline
M3D-RPN \cite{brazil2019m3d} &50.24&- &40.01 &-&\textbf{33.48}  & - &20.45&- &17.03 &-&15.32 & -\\
+RAR-Net&\textbf{51.20}&- &\textbf{44.12} &-&32.12 & - &\textbf{23.12}&- &\textbf{19.82} &-&\textbf{16.19 }& -\\
\hline
\end{tabular}
}
\end{center}
\label{tab:location}
\end{table}

\begin{table}[t]
\caption{3D detection accuracy (AP, \%) in the KITTI test set (in each group, the left accuracy is produced by M3D-RPN, and the right one by M3D-RPN+RAR-Net).}
\begin{center}
\begin{tabular}{|c|c|c|c|}
\hline
Metirc &  Easy & Moderate& Hard \\
\hline\hline
AOS &	   88.38/88.48 & 82.81/83.29 & 67.08/67.54\\
\hline
Bird &	21.02/22.45 & 13.67/15.02 & 10.23/12.93\\
\hline
3D AP & 14.76/16.37 & 9.71/11.01 & 7.42/9.52\\
\hline
\end{tabular}
\end{center}
\label{tab:dimension2}
\end{table}

Table \ref{tab:dimension2} shows our results on the KITTI test set using M3D-RPN as baseline, which is consistent with the results in the validation set. We also tried to use D4LCN\cite{ding2020learning} as a baseline, which used additional depth data for training, and we can still observe accuracy gain (0.51\% AP) with a smaller step size (0.02).

\subsection{Diagnostic Studies} In the ablation study we want to analyze the contributions of different sub-module and different design choices of our framework. In Table \ref{tab:ablation}, We use the initial detection results of MonoGRNet~\cite{qin2019monogrnet} as baseline. \textsf{Discrete Output} is to output a discrete refining choice instead of a continuous refining value. We also tried three different feature combining methods: \textsf{Simple Fusion} is the naive option which concatenates the current detection results parameters and the image feature vector, \textsf{Direct Projection} is to project the bounding box on the original image as \cite{liu2019deep} did, and \textsf{Parameter-aware} means our parameter-aware module. We refer \textsf{Axial Coordinate} to the option of refining the location along the axial coordinate system rather than the world coordinate system. \textsf{Single Action} is to output one single refinement operation at a time rather than output all refinement operations for all the 3D parameters at the same time. \textsf{RL} is to optimize the model using RL. \textsf{Final Model} is our fully model with the best design choices.

\begin{table}[t]
\caption{Ablation experiments on KITTI dataset (val 1, Easy, IoU = 0.7). The performance difference can be seen by comparing each column with the last column.}
\small
\renewcommand\tabcolsep{3.5pt}
\label{table:ablation}
\centering
\begin{tabular}{|l|c|c|c|c|c|c|c|}
\hline
\multirow{1}*{Module} & \multicolumn{6}{c|}{Design Choices} & \multicolumn{1}{c|}{\textsf{Final Model}} \\
\hline
\textsf{Discrete Output}&  &  \checkmark  &\checkmark & \checkmark & \checkmark & \checkmark& \checkmark\\
\hline
\textsf{Simple Fusion} &   & \checkmark &   &  &  &  &\\
\textsf{Direct Projection}&    &  & \checkmark &&   &  &\\
\textsf{Parameter-aware}&  \checkmark &  &  & \checkmark&  \checkmark & \checkmark& \checkmark\\
\hline
\textsf{Axial Coordinate}&  \checkmark & \checkmark& \checkmark&  &\checkmark &  \checkmark& \checkmark \\
\hline
\textsf{Single Action }&   \checkmark& \checkmark&  \checkmark& \checkmark&  & \checkmark &\checkmark\\
\hline
\textsf{RL} &  \checkmark& \checkmark& \checkmark & \checkmark&  & &\checkmark\\
\hline
3D AP & 1.81 & 0.40 & 10.88  & 5.34 &  2.27 &13.96  &18.25 \\
\hline
\end{tabular}
\label{tab:ablation}
\end{table}

Through comparing \textsf{Discrete Output} with \textsf{Final Model}, we find that directly regressing the continuous 3D parameters can easily lead to a failure in refinement and with controlled discrete refinement stride, the results can be much better. Also, we can see that \textsf{Simple Fusion} does not work well, which verifies that our image enhancement approach captures richer information. Besides, moving along the axial coordinate system and using the single refinement operation can also improve the performance and verify our arguments. Experiment also demonstrate that RL play an important role in boosting the performance further since it optimizes the whole refinement process.

We notice that the number of steps and the refining stride have great impact to the final refinement results. So, during the test phase, we have tried different setting of steps and stride. With smaller strides and more steps, better performance can be achieved but with lager time cost. In addition, when the strides are too large, the initial 3D box of an object may jump to its neighboring object occasionally and some false positives can also be adjusted to overlap with an existing, true 3D box by accident. Since the moving stride and steps are also a part of the refinement policy, using RL to optimize them is feasible as well.

\begin{figure}[t]
\begin{center}
    \begin{center}
       \includegraphics[width=1\linewidth]{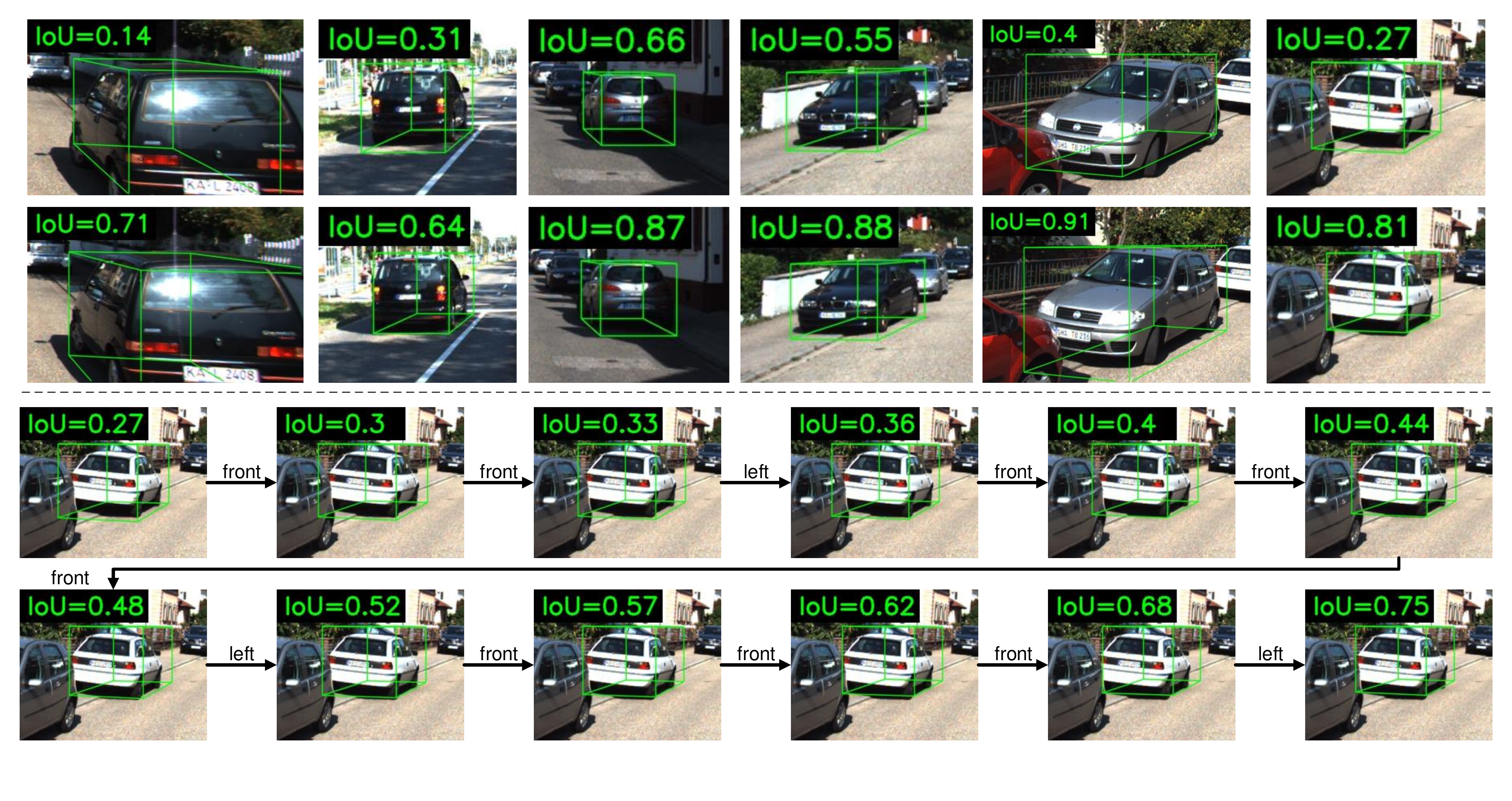}
    \end{center}
\end{center}
   \caption{Top 2 rows: Representative examples on which the proposed refinement method achieves significant improvement beyond the baseline detection results. The rightmost example is further detailed in the bottom 2 rows. }
\label{fig:qualitative}
\end{figure}

Last but not least, we visualize some refinement results in Fig. \ref{fig:qualitative}, where the initial 3D bounding box and the final refinement result are in shown with their 3D-IoU to ground-truth. We can see that our refinement method can refine the 3D bounding box from a coarse estimate to the destination where it can fit the object tightly. Apart from drawing the starting point and ending point of 3D detection boxes on 2D images, we also show some intermediate result for better understanding. During each iteration, our approach can output a refining operation to increase the detection performance.

\subsection{Computational Costs}
We also compute the latency for our model. Our method achieves about 4\% improvement compared to baseline, with a computation burden of 0.3s (10 steps), which is much smaller than the detection time cost: 2s (GS3D \cite{li2019gs3d}). Generally speaking, the cost is related to three aspects: (1) network backbone, (2) number of steps, (3) number of objects. For (1), using smaller backbone (such as ResNet-18) can further speed up the refinement process with some degraded performance. For (2), we can increase the refining stride of each step that will cause the number of steps to drop and further accelerate the refining stage, with the price of some imperfect correction. For (3), multiple objects in one image can be fed into the GPU as a batch and processed in parallel, so the inference time does not increase significantly compared to a single object. 

\section{Conclusions}
In this paper, we have proposed a unified refinement framework called RAR-Net. In order to use multi-step refinement to increase the sampling efficiency, we formulate the entire refinement process as an MDP and use RL to optimize the model. At each step, to fuse two information sources from the image and 3D spaces into the same input, we project the current detection into the image space, which maximally preserves information and eases model design.  quantitative and qualitative results demonstrate that our approach boost the performance of the state-of-the-art monocular 3D detectors with a small time cost. 

The success of our approach sheds light on applying indirect optimization to improve the data sampling efficiency in challenging vision problems. 
We believe that inferring 3D parameters from 2D cues will be a promising direction of a variety of challenges in the future research.

\subsubsection*{Acknowlegements}
This work was supported in part by the National Key Research and Development Program of China under Grant 2017YFA0700802, in part by the National Natural Science Foundation of China under Grant 61822603, Grant U1813218, Grant U1713214, and Grant 61672306, in part by Beijing Natural Science Foundation under Grant No. L172051, in part by Beijing Academy of Artificial Intelligence (BAAI), in part by a grant from the Institute for Guo Qiang, Tsinghua University, in part by the Shenzhen Fundamental Research Fund (Subject Arrangement) under Grant JCYJ20170412170602564, and in part by Tsinghua University Initiative Scientific Research Program.

\bibliographystyle{splncs04}
\bibliography{egbib}
\end{document}